\documentclass{article}

\usepackage[preprint]{paper/neurips/neurips_2026}
\usepackage{amsmath}
\usepackage{amsthm}
\newtheorem{definition}{Definition}
\usepackage{graphicx}
\usepackage{algorithm}
\usepackage{algpseudocode}
\usepackage{comment}
\usepackage[utf8]{inputenc} 
\usepackage[T1]{fontenc}    
\usepackage{hyperref}       
\usepackage{url}            
\usepackage{booktabs}       
\usepackage{amsfonts}       
\usepackage{nicefrac}       
\usepackage{microtype}      
\usepackage{xcolor}         
\usepackage{float}
\usepackage{wrapfig}
\usepackage{placeins}
\usepackage{enumitem}

\title{scShapeBench: Discovering geometry from high dimensional scRNAseq data}

\author{
Andrew J. Steindl$^{1}$\thanks{Equal contribution.}
\And João Felipe Rocha$^{1}$\footnotemark[1]
\And Brian Tshilengi Di Bassinga$^{1}$
\And Zachary Warren$^{1}$
\AND
Matthew Scicluna$^{2}$
\And César Miguel Valdez Córdova$^{2}$
\And Shabarni Gupta$^{3}$
\And Leire Torices$^{3}$
\AND
Daniel Neumann$^{4}$
\And Timothy J. Mann$^{4}$
\And Ihuan Gunawan$^{4}$
\And Dhananjay Bhaskar$^{5}$
\AND
John G. Lock$^{4}$
\And Christine L. Chaffer$^{3}$
\And Guy Wolf$^{2}$
\And Smita Krishnaswamy$^{1}$\thanks{Corresponding author.}
\\[0.75em]
$^{1}$Yale University
\qquad
$^{2}$Mila / Université de Montréal
\\
$^{3}$Garvan Institute of Medical Research
\qquad
$^{4}$School of Biomedical Sciences, University of New South Wales
\\
$^{5}$University of Wisconsin--Madison
\\[0.5em]
\texttt{drew.steindl@yale.edu}
\quad
\texttt{joaofelipe.rocha@yale.edu}
\quad
\texttt{smita.krishnaswamy@yale.edu}
}

\begin{document}

\maketitle

\begin{abstract}
High-dimensional point cloud data arise across many scientific domains, notably single-cell biology. The "shapes" or topologies of these datasets are informative of types of information that can be extracted from the datasets. For example clustered data admits the extraction of cell types or cell states in a static analysis of the datasets. Continuous trajectory structures admit continuous transition or trajectory analysis, while other shapes such as archetypal shapes admit continuum extraction with a range of cells spanning behaviors. While analysis pipelines exist, they often presuppose shape in data. For example, the standard Seurat pipeline combines UMAP visualization with Louvain clustering. This assumes clustered data. Tools like Monocle and Spade assume a tree-like shape, and flow-models like MIOFlow and Conditional Flow Matching are suitable for trajectories. Deciding which pipeline to apply to which part of the data is often the realm of bioinformaticians who visualize and qualitatively analyze the data before selecting one. However, with the advent of agentic AI scientists, it becomes important to automate data shape detection, particularly into categories that are relevant for downstream analysis pipelines. Towards this end we introduce \textsc{scShapeBench} a benchmark dataset comprising both synthetic and single-cell expert-annotated datasets that are meant for the task of shape detection. Synthetic datasets are sampled from a ground truth “skeleton graph” with variance. Real single-cell datasets are curated from a variety of sources and are annotated by experts, classifying four categories; clusters, single trajectory, multi-branches and archetypes. In addition, we provide a baseline method, scReebTower, to bridge the gap between data visualization and pipeline selection. scReebTower relies on the diffusion geometry to extract Reeb graphs. We provide new topology-aware metrics with which we evaluate scReebTower and existing methods PAGA and Mapper on synthetic data. On single-cell data we curate expert annotations of shapes, and showcase evaluations of methods. Our comparisons indicate scReebTower outperforming other baselines. Overall, our contributions span benchmarks, evaluation metrics, and a novel baseline method for automated shape detection in high-dimensional single-cell data.
\end{abstract}

\section{Introduction}

High-dimensional datasets arising in biology, neuroscience, and machine learning often possess intrinsic low-dimensional structure that is critical to their interpretation. A researcher analyzing such a dataset faces an immediate question: does the data organize into discrete cell types, a continuous differentiation trajectory, a cyclic process or some combination? In short: what \textit{shape} does the data have and what tools extract information accordingly?

\begin{figure}[t]
    \centering
    \includegraphics[width=1\linewidth]{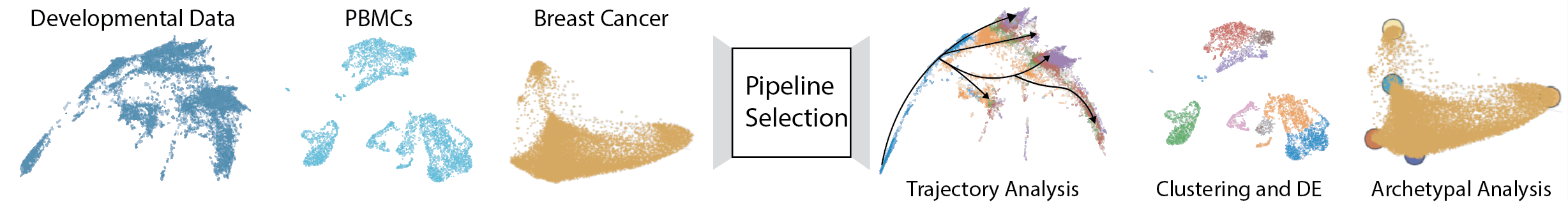}
    \caption{Pipeline selection is a common challenge in single-cell data analysis. While appropriate methods exist, unsupervised selection of downstream analysis based on point cloud structure remains a problem that has not been widely addressed.}
    \label{fig:pipeline}
    \vspace{-.6cm}
\end{figure}

This question matters because specialized analysis methods exist for each shape, and matching method to shape is essential for recovering biological signal (Figure \ref{fig:pipeline}). The standard Seurat \citep{seurat} pipeline assumes cluster structure motivated by measurements of fully differentiated cell types such as B cells and T cells in circulating blood cells (PBMCs).  Monocle 3 \citep{monocle3} and SPADE \cite{anchang2016visualization} are motivated by branching processes such as hematopoiesis, or when treatment response creates a bifurcation in response. MIOFlow \citep{mioflow} and TrajectoryNet \citep{tong2020trajectorynet} assume continuous trajectories that are found in transitions from primary to metastases or other disease-related transformations. While these are the motivations for these methods, just knowing a measurement is based on blood cells does not necessarily imply cluster-structure it closely depends on the processes active in the cells during the conditions measured. For instance, are the blood cells responding to treatment, or a viral infection? The final decision is often best informed by the data itself. 

Yet shape cannot be read off the raw high-dimensional data easily. While many tools exist to process the geometries of the data into visualizations, there is a lack of tools that extract a simple explanation of the shape of the data. Currently shape detection is a manual process: a bioinformatician inspects a low-dimensional embedding and selects methods through a mix of domain expertise and visual intuition. In absence of such intuition,  default choice is almost always the Seurat \citep{seurat} clustering pipeline, regardless of whether the data is actually cluster-structured, and real biological signals encoded in trajectories, cycles, or hybrid geometries are routinely missed as a result.

This manual step is a bottleneck for any automated analysis system. Automating this selection requires a method capable of detecting intrinsic {\em shape} or topology of the data, in a quantitative and mathematically principled fashion. While some methods, such as PAGA ~\citep{PAGA}, or Mapper ~\citep{MapperPaper} aim to provide simplified representations of the data, their effectiveness in helping automate shape detection in single-cell data has not been systematically tested. Indeed, to our knowledge this problem has not been formulated mathematically, and evaluation metrics do not exist for this purpose. 

We define the problem of {\em single-cell shape detection} as the problem of recovering a graph  \(S\) from a high-dimensional point cloud \(X\), such that \(S\) reflects the underlying topology of the data (see Figure \ref{fig:shape-recovery}). To support work on this problem, we introduce \textsc{scShapeBench}, a benchmarking framework comprising of:

\begin{enumerate}
    \item \textbf{A synthetic dataset with ground-truth topology.} We provide synthetic point clouds generated from ground-truth graphs with known topology across four shape classes with controlled variation in noise, dimensionality, and sampling density. These enable exact evaluation of topological correctness via direct comparison to the generating graph.

    \item \textbf{An expert-annotated scRNAseq corpus.} We curate a collection of 102 real-world scRNAseq datasets for which 9 expert biologists and bioinformaticians have annotated the shape. These enable evaluation of shape recovery where ground-truth topology is unavailable but annotations provides a reliable reference. The full annotated dataset is released with structured datasheets hosted on Hugging Face.
 
    \item \textbf{Topology-aware evaluation metrics.} We defined metrics for comparing recovered graph skeletons to reference structures including graph edit distance, and persistence similarity. For real-world datasets lacking a ground-truth graph, we evaluate whether recovered skeletons match the topological class annotated by domain experts.
 
    \item \textbf{A standardized evaluation protocol.} We define a consistent evaluation interface (point cloud in, graph out), apply uniform preprocessing, compare a set of existing methods across all shapes and released a reproducible evaluation pipeline for any new developed method.

    \item \textbf{A baseline method.} We introduce \textbf{scReebTower}, which constructs Reeb graphs \citep{Reeb1946} from diffusion maps. \textbf{scReebTower} achieves state-of-the-art performance on \textsc{scShapeBench} and improves over baselines demonstrating that the benchmark is not merely diagnostic but enables development of better methods.
\end{enumerate}

\section{Background}
\vspace{-.3cm}
\label{sec:background}
In this section we define and explain concepts from graph theory,  diffusion geometry and topological data analysis that are pertinent to the problem of single-cell shape detection.
\vspace{-.3cm}
\subsection{Graph theory}
\vspace{-.3cm}
\noindent\textbf{Graphs.} A graph \(G=(V,E)\) consists of a set of vertices \(V\) together with a set of edges \(E\), where each edge connects a pair of vertices. Since \(E\) is a set, duplicate edges are not permitted: there is at most one edge between any pair of vertices. Such graphs are commonly referred to as \emph{simple graphs}. A \textbf{\textit{multigraph}} generalizes a \textit{simple graph} by replacing the edge set with a multiset, allowing multiple edges between the same pair of vertices.

\noindent\textbf{Reduced graphs.}\label{par:reducedgraphs}
Infinitely many graphs differ only by the presence of degree-two nodes along paths. Since these nodes do not change the overall branching or connectivity structure of the graph, it is often convenient to remove them. For a graph \(G\), let \(\widetilde{G}\) denote the \emph{reduced graph} obtained by contracting maximal paths whose internal vertices have degree \(2\). This preserves branch points, pendant edges, cycles, and connected components while removing redundant vertices.

\subsection{Diffusion Geometry}
\vspace{-.3cm}
Diffusion geometry studies the intrinsic geometry of data through diffusion processes, such as heat propagation, on graphs constructed from the data. This was introduced in the context of manifold learning with diffusion maps (DMs) \citep{coifman2006diffusion}.

\noindent\textbf{Kernels.}
A kernel transforms pairwise distances between data points ($x_n$) into affinities, producing a weighted graph in which nearby points receive large weights and distant points receive small weights. Throughout this work we use the adaptive Gaussian kernel
\begin{equation}
w_{ij} = \exp\!\left(-\frac{\|x_i-x_j\|^2}{\sigma_i \sigma_j}\right),
\label{eq:kernel}
\end{equation}
where \(\sigma_i\) is a local scale parameter that adapts to the density of the data around point \(x_i\). Collecting these affinities into a matrix \(W\) with entries \([W]_{ij} = w_{ij}\) yields the \emph{affinity matrix} of the data.

\noindent\textbf{Diffusion operator.}
Row-normalizing \(W\) yields a row-stochastic matrix \(P\), the \emph{diffusion operator},
\begin{equation}
\label{eq:dif_op}
    P = D^{-1} W
\end{equation}
where \(D\) is the diagonal degree matrix with \(D_{ii} = \sum_j W_{ij}\). The entry \(P_{ij}\) is the probability of transitioning from \(x_i\) to \(x_j\) in a single step of a Markovian random walk on the data graph. Powers \(P^t\) encode \(t\)-step transition probabilities, $P^t_{ij} = \Pr(x_t = x_j \mid x_0 = x_i),$ with each row \(P^t_{i,:}\) giving the \(t\)-step diffusion distribution starting from \(x_i\).

\noindent\textbf{Diffusion maps and eigenvectors}
The \(L^2\) distance between rows of \(P^t\) defines a \emph{diffusion distance}, measuring similarity through graph diffusion rather than ambient geometry. The spectral decomposition of \(P\) realizes this distance geometrically, with eigenvectors \(\phi_0, \phi_1, \ldots, \phi_n\) ordered by eigenvalues $1 = |\lambda_0| \geq |\lambda_1| \geq |\lambda_2| \geq \cdots .$ These eigenvectors act as frequency harmonics of increasing order. Since \(P\) is a Markov matrix, \(\lambda_0 = 1\); for connected graphs, the corresponding leading right eigenvector is constant, while multiplicity of the eigenvalue \(1\) reflects disconnected components. The first non-trivial diffusion eigenvector \(\phi_1\), the Fiedler vector, is the lowest-frequency nonconstant mode and varies smoothly across the graph. In this work, its level sets reveal the underlying shape of the point cloud.

\noindent\textbf{Diffusion condensation}
Diffusion condensation is a methodology developed in \cite{smitaDiffusion, smitaDiffusion2} which presents a method for systematic coarse graining of data via application of the diffusion operator back to the data. At each iteration this replaces each original data vector \(X_i\) with a weighted average of its $t$-step diffusion neighbors: \(X^{(t)} = P^t X\). Then a new diffusion operator is computed from the condensed datapoints $X^{(t)}$. This process is proven in \cite{smitaDiffusion} to eventually converge to a single point thereby sweeping all granularities of data. It therefore serves as a continuously hierarchical clustering method. We refer to the number of iterations of this process as $\ell$.

\subsection{Topological Data Analysis}
\vspace{-.3cm}
Topological Data Analysis (TDA) studies the shape of data using tools from algebraic topology, with the goal of recovering features such as connected components, loops, and branches.

\begin{wrapfigure}{r}{0.6\textwidth}
    \centering
    \includegraphics[width=.8\linewidth]{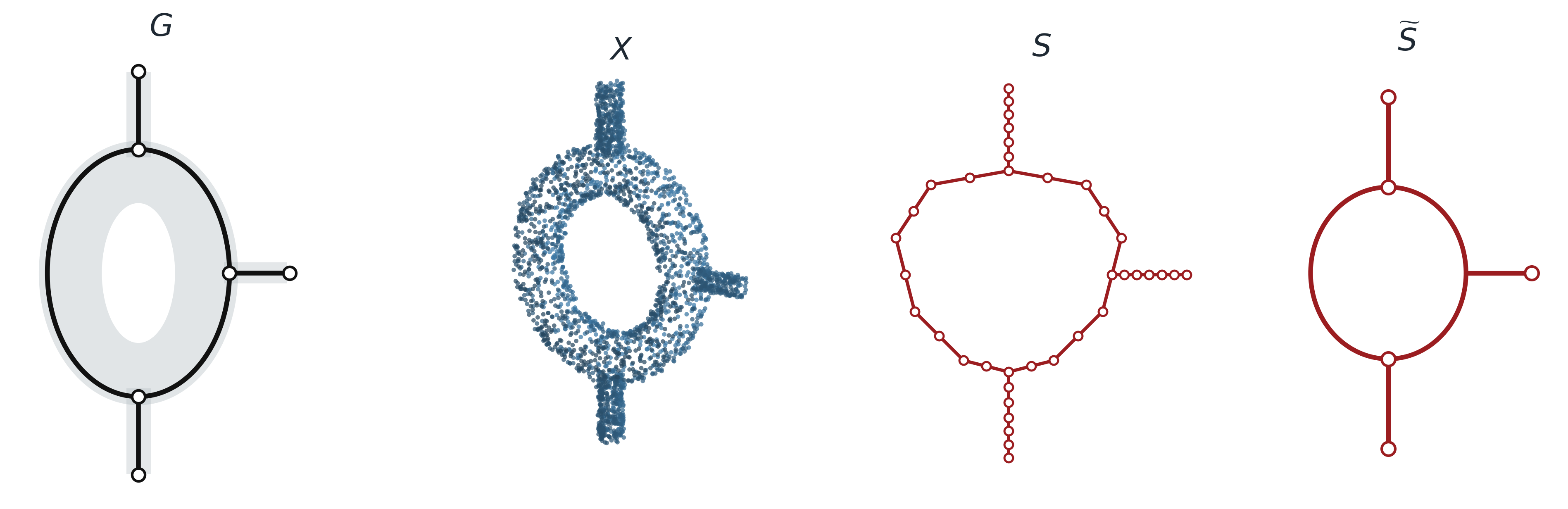}
\caption{\textbf{Single-cell shape recovery framework.}
The goal of shape recovery is to infer a graph representation of the underlying organization of high-dimensional data. \textbf{(a)} A latent graph \(G\) describes the underlying organization of the data. \textbf{(b)} Samples from this structure produce the observed noisy high-dimensional point cloud \(X\). \textbf{(c)} A shape recovery method constructs a graph \(S\) from the observed data, potentially introducing additional degree-two vertices along paths. \textbf{(d)} Graphs are reduced by suppressing degree-two vertices, yielding the reduced graph \(\widetilde{S}\).}    \label{fig:shape-recovery}
\end{wrapfigure}

\noindent\textbf{Persistent homology.}
Persistent homology tracks how topological features such as connected components and cycles appear and disappear as a scale parameter varies, providing a multi-scale summary of shape. One constructs a \emph{filtration}—a growing sequence of spaces—and records the (birth, death) scale of each feature, typically visualized as a \emph{persistence diagram} or barcode. Long-lived features are interpreted as genuine structure, while short-lived ones are attributed to noise.
For graphs, a natural choice is the \emph{edge-length filtration}: edges are added from shortest to longest, merging components (deaths) and forming cycles (births). The triggering edge lengths give the birth and death scales, distinguishing prominent loops and clusters from spurious ones.

\noindent\textbf{Morse functions.}
A Morse function is a smooth scalar-valued function on a space whose critical points are well-defined and isolated.

\noindent\textbf{Reeb graphs.}
A Reeb graph summarizes how the connected components of level sets of a Morse function evolve over a space. Connected components of level sets are represented as nodes, while edges track how these components merge or split as one sweeps through the function values.

Our main challenge is to recover a reduced graph from noisy data, and the proposed method \textbf{scReebTower} uses diffusion geometry to uncover this structure. We discuss related work on similar problems in Appendix~\ref{app:related_works}.

\vspace{-.3cm}
\section{Problem Formulation}
\label{sec:problem-formulation}
\vspace{-.3cm}

Let \(X \in \mathbb{R}^{n \times m}\) be a high-dimensional point cloud such as a cell-by-gene matrix arising from scRNAseq data. We view \(X\) as a noisy, non-uniform sample from a latent geometric structure admitting a latent graph representation \(G\).

For the purposes of this work, we define the \emph{shape} of a dataset to be the reduced graph representation of its latent organization: the connected components, branches, cycles, and endpoints that remain after suppressing degree-two subdivision vertices.

\begin{definition}[\textbf{Single-cell shape detection}]
Given a point cloud \(X\) sampled from a latent geometric structure admitting a latent graph representation \(G\), the \emph{single-cell shape detection problem} is to recover a graph \(S\) that optimizes
\[
\mathcal{C}(\widetilde{S}, \widetilde{G}),
\]
where \(\widetilde{S}\) and \(\widetilde{G}\) denote the reduced graphs of \(S\) and \(G\), and \(\mathcal{C}(\cdot,\cdot)\) is a topology-aware comparison function, maximized or minimized depending on the chosen metric.
\end{definition}

Evaluating this objective requires datasets in which the target shape is accessible. In our synthetic benchmark, the latent graph \(G\) is known by construction, enabling direct comparison between \(\widetilde{S}\) and \(\widetilde{G}\) via \(\mathcal{C}\) instantiated through edge-length persistent homology and graph edit distance. In the scRNAseq benchmark, where no ground-truth graph exists, recovered graphs are instead evaluated through downstream prediction of expert-annotated shape labels from graph-derived features.

These requirements motivate the construction of \textsc{scShapeBench}, which collects datasets where either the latent graph is known exactly or the shape is sufficiently characterized to support evaluation.

\vspace{-.3cm}
\section{\textsc{scShapeBench} datasets}
\label{sec:datasets}
\vspace{-.3cm}
Evaluating shape recovery requires datasets with known or well-characterized shape. This is particularly challenging for scRNAseq data, where the underlying organization is not directly observed. We therefore construct \textsc{scShapeBench} from two complementary sources: \textbf{(i) synthetic datasets} generated from predefined latent graphs, where the underlying shape is known exactly by construction, and \textbf{(ii) an expert-annotated scRNAseq corpus}, where shape is inferred from biological knowledge and expert annotation. The relationship between these settings is illustrated in Figure~\ref{fig:shape-recovery}.

\subsection{Synthetic data}
\begin{wrapfigure}{r}{0.6\textwidth}
    \centering
    \vspace{-10pt}
    \includegraphics[width=0.6\textwidth]{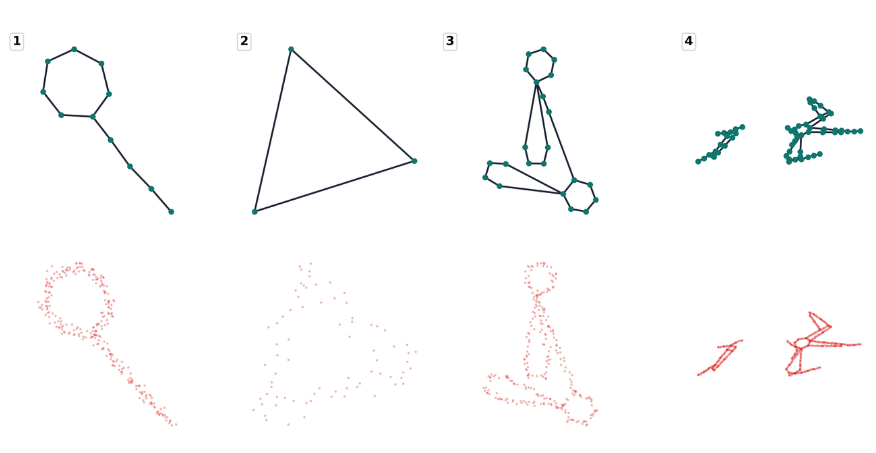}
    \caption{Representative synthetic benchmark exemplars. The top row shows the 2D latent ground-truth graphs and the bottom row shows the corresponding sampled point clouds. The examples span increasing complexity from a simple shape to an archetypal triangle, a graph with multiple cycles, and a difficult disconnected example.}
    \label{fig:synthetic_examples}
    \vspace{-10pt}
\end{wrapfigure}

\vspace{-.3cm}
Following the formulation in Section~\ref{sec:problem-formulation}, we generate noisy point clouds \(X\) from embedded latent graphs \(G\). The recovery objective is to construct a graph \(S\) whose reduced form \(\widetilde{S}\) matches the reduced latent graph \(\widetilde{G}\). The benchmark spans trees, cycles, and hybrids thereof, varied across four difficulty axes: noise scale, feature separation, sampling density, and feature thickness (Figure~\ref{fig:synthetic_examples}).

\noindent\textbf{Graph generation.}
Graphs are generated by sampling one or more connected components independently. Each component is sampled from six possible graph classes:
singletons, single edges, trees,
single cycles, multiple cycles, and hybrids.
For components containing cycles, we first generate a cycle backbone by assembling cycles into separated or fused configurations. Acyclic branches are then attached recursively to eligible vertices. Additional connected components are added with decreasing probability and placed into a shared ambient space with controlled separation, producing disconnected examples when appropriate.

\noindent\textbf{Embedding, sampling, and difficulty axes.}
Given a sampled graph, we construct an embedding together with a corresponding point cloud using the pipeline summarized in Algorithm~\ref{alg:graph_generation}. Graphs are embedded in low dimension, typically \(\mathbb{R}^2\), with higher ambient dimensions used when necessary to realize more complex structures. Edges are then thickened into tubular geometric features, and separation transformations are applied to produce both well-separated and densely packed configurations. Noisy point clouds are sampled around the resulting embedded graph structure with additive Gaussian noise. Benchmark difficulty is controlled along four axes: noise scale, feature separation, sampling density, and feature thickness. These parameters are sampled independently across preset ranges so that the resulting dataset spans both easy and difficult shape recovery regimes. Each synthetic sample is saved together with its generating graph \(G\).

\subsection{Single-cell RNA sequencing data}
\vspace{-.3cm}
We curated a collection of 102 real-world scRNAseq datasets annotated by 9 expert biologists and bioinformaticians. Because scRNAseq data has no known ground-truth shape, recovered graphs are evaluated against these expert annotations. The curation procedure consisted of three stages: 1) assembling a diverse pool of datasets from public repositories, 2) applying a standardized preprocessing pipeline together with two visualizations to support annotation and analysis, and 3) collecting expert shape annotations.

\noindent\textbf{Dataset selection}
The single-cell data corpus comprises $102$ scRNAseq datasets manually curated from four public repositories:
(i) The CELLxGENE Census~\citep{czi2025cz} hosted by the Chan Zuckerberg Initiative ($62$ datasets), (ii) The public catalog of $10$x~Genomics ($30$ datasets), (iii) The Broad Institute Single Cell Portal ~\citep{tarhan2023single} ($8$ datasets), (iv) The EMBL-EBI Single Cell Atlas~\citep{george2024expression} ($1$ dataset).

We opted for manual curation across multiple sources rather than a broad automated scrape for two reasons. First, a large fraction of publicly available scRNAseq datasets are noise-dominated measurements exhibiting approximately Gaussian geometry, providing little meaningful shape information and therefore dominating any uncurated collection. We therefore filtered such datasets out. Second, no single repository spans all major shape classes on its own: the \(10\times\) catalog skews toward cluster-like datasets (e.g., PBMCs), while CELLxGENE and the Broad SCP contribute differentiation and disease studies with more branching and trajectory structure. The resulting per-dataset metadata, organism, tissue, assay coverage, and cell-count distributions are reported in Appendix~\ref{app:corpus}.

\noindent\textbf{Preprocessing pipeline.}
All datasets are processed through a single standardized Scanpy~\citep{wolf2018scanpy} pipeline including filtering, log-normalization, and subsampling to \(50{,}000\) cells. This standardization ensures that downstream methods operate on directly comparable inputs. The fixed cell count is deliberate: many graph-based methods require hyperparameter choices that depend strongly on sampling density and dataset size (e.g., neighborhood sizes in \(k\)-NN graph construction). Using a common sample size therefore improves the stability and comparability of method outputs across datasets while keeping memory and runtime costs tractable as the benchmark scales. Full configuration details are provided in Appendix~\ref{app:preprocessing}.

\noindent\textbf{Expert annotation}

\label{sec:expert-annotation}
\begin{wrapfigure}{r}{0.5\textwidth}
    \centering
    \vspace{-10pt}
    \includegraphics[width=\linewidth]{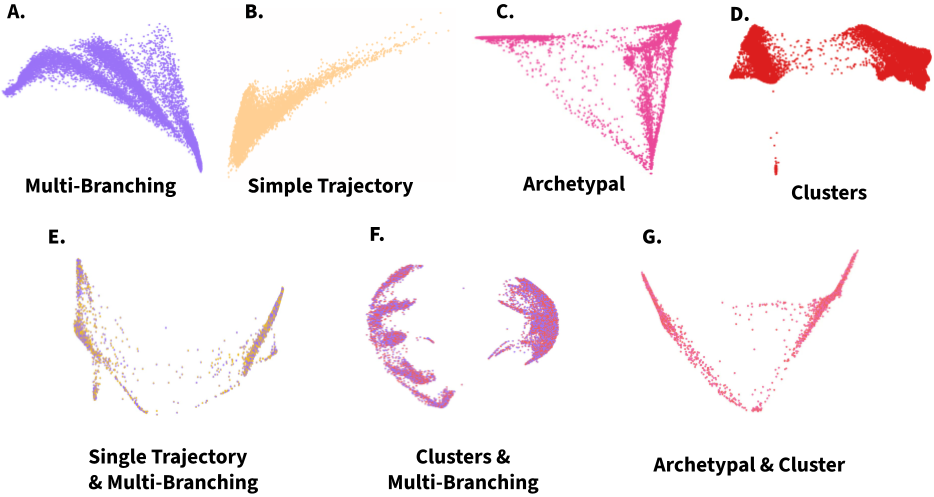}
    \caption{PHATE visualizations of experts annotations. Some datasets can be classified as more than one category.}
    \label{fig:real_datasets}
    \vspace{-10pt}
\end{wrapfigure}

Because expert annotation is performed visually, we computed 2-dimensional PHATE \citep{phate} and UMAP \citep{umap} embeddings of each preprocessed dataset. We chose this pair because PHATE is designed to preserve both local and global manifold structure and has been shown to retain geometry and preserve trajectories and cycles \citep{phate,kuchroo2022multiscale}, while UMAP provides a cluster-centric view that is familiar to biologists. Through a webpage gallery (see Appendix \ref{app:webpage}), nine annotators viewed these embeddings and assigned each dataset shape labels drawn from  $\mathcal{L} = \{\textsc{clusters}, \textsc{single-trajectory},  \textsc{multi-branching}, \textsc{archetypal}\}$. While this label set can be extended to additional shapes, we restricted it to ensure consistent annotation across annotators. Since scRNAseq datasets frequently exhibit more than one organizational regime simultaneously (i.e. distinct cell types progressing  through the cell cycle) annotators may select any non-empty subset of the labels. The shapes are defined by the following visual cues and biological motivations (examples can be seen on Figure \ref{fig:real_datasets}):

\begin{itemize}[leftmargin=*]
    \item \textsc{Clusters}: discrete, well-separated point groups with visible empty space between them in the 2D embedding. Produced when measured cells correspond to terminally differentiated, stable cell types (e.g. tissue-resident immune populations \cite{milner2020heterogenous}) [Figure \ref{fig:real_datasets}D].
    \item \textsc{Single-trajectory}: an one-dimensional ribbon of points without bifurcations. Produced by linear biological processes such as a synchronized stimulus response over time\citep{krishnaswamy2023revealing} [Figure \ref{fig:real_datasets}B].
    \item \textsc{Multi-branching}: a tree-like structure exhibiting one or more bifurcation points at which a single arm splits into two or more downstream branches. Produced by developmental processes with explicit cell-fate decisions, such as hematopoiesis, neurogenesis, or organoid differentiation \citep{phate} [Figure \ref{fig:real_datasets}A].
    
    \item \textsc{Archetypal}: a polygonal or simplex-shaped envelope with extreme specialist cells occupying the vertices and intermediate cells filling the convex interior. Produced when cells balance multiple specialized functions, such as cancer cells balancing proliferation, hypoxic adaptation and adipogenic metabolism \citep{venkat2025aanet} [Figure \ref{fig:real_datasets}C].
\end{itemize}
\vspace{-.3cm}

\noindent\textbf{Aggregation of annotations.}
\textsc{scShapeBench} is designed to evaluate the full set of methods a biologist might apply to a dataset. We therefore treat any method class selected by at least one annotator as part of the relevant tool set for that dataset: a label is marked positive if any annotator selected it. A stricter majority rule would discard exactly the cases where methodological pluralism matters most, datasets on which reasonable experts disagree about the dominant downstream analyses. Under union aggregation, all nine annotators contribute to every retained dataset's labels, and all four shape classes are preserved in the primary evaluation. To guard against low-confidence annotations propagating into ground truth, we report sensitivity analyses in Appendix~\ref{app:aggregation}. 

\noindent\textbf{Release format.}
All datasets in \textsc{scShapeBench} are derived from previously published or vendor-released public scRNAseq resources, and we redistribute them under their original licenses, recorded per-dataset in the accompanying datasheet. Each preprocessed dataset is released as a single \texttt{h5ad} file containing the normalized log-counts in \texttt{.X}, the raw integer counts in \texttt{.raw}, all original metadata fields inherited from the source repository, and the enriched bookkeeping fields described above. The expert structural labels and the $2$D PHATE coordinates used for annotation are released alongside the corpus as a separate manifest keyed by \texttt{dataset\_id}. The full corpus, datasheet, raw and aggregated labels are hosted on Hugging Face at \url{https://huggingface.co/datasets/scShape-Benchmark/scShapeBench}. No personally identifying information is included in the released benchmark. The scRNAseq dataset is accompanied by a datasheet containing provenance, preprocessing details, licensing information, annotation statistics, and release metadata.

\section{The scReebTower Method}

\begin{wrapfigure}{r}{0.55\textwidth}
    \centering
    \vspace{-10pt}
    \includegraphics[width=0.53\textwidth]{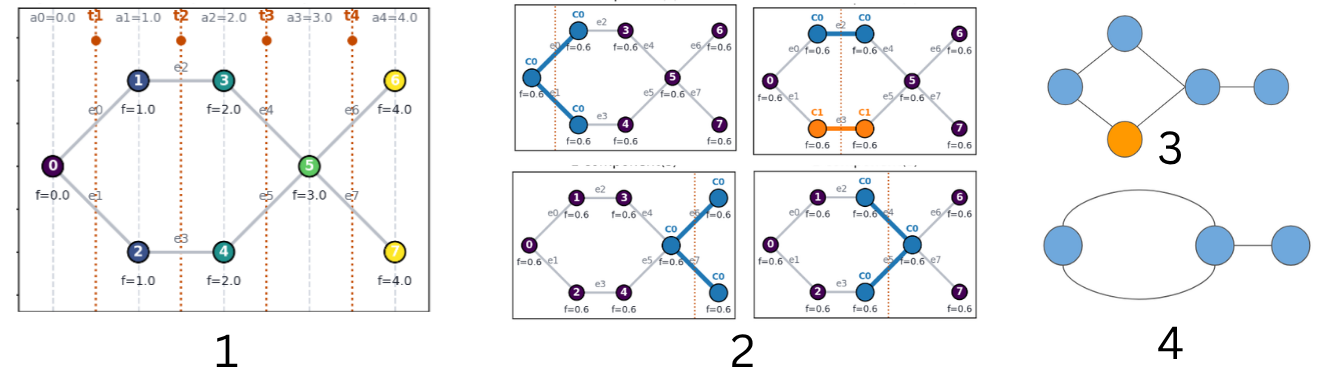}
    \caption{Overview of the \textsc{scReebTower} algorithm. (1) Data colored by the discrete Morse function, with midpoint thresholds indicated. (2) Level sets defined by edges crossing each threshold. (3) The resulting Reeb graph. (4) The simplified Reeb graph after suppressing degree-\(2\) subdivision vertices while preserving cycles.}
    \label{fig:reebalg}
    \vspace{0pt}
\end{wrapfigure}
We provide a simple, but powerful method for identifying the data shape, from which the label can be detected, called \textbf{scReebTower} (Figure \ref{fig:reebalg}).  Given a dataset \(X\), \textbf{scReebTower} first constructs a \(k\)-nearest-neighbor graph, on $X$ and computes the diffusion operator $P$ from an adaptive Gaussian kernel as described in Equation \ref{eq:dif_op}. The Fiedler vector $f(x)=\phi_1(x)$, is then used as a discrete Morse function, $f$ on the data defined on each node $f(x)$. Level sets of $f^{-1}(x)$ are tracked across the \(k\)-nearest-neighbor graph, and within each level the connected components are contracted to single nodes. Nodes in adjacent levels are linked to form a Reeb graph \(S\), which is subsequently reduced to \(\widetilde{S}\).

Higher levels of \textbf{scReebTower} are constructed iteratively by applying the diffusion condensation operation \cite{smitaDiffusion2} described in Section \ref{sec:background}. This condensation proceeds iteratively by replacing each datapoint with a weighted average of its diffusion neighbors $\hat{X}=P^tX$, and a new diffusion operator is constructed for this level $\hat{P}$. Multiple iterations create a non-homogenous Markov process that gradually coarse grains the graph. A simplified Reeb graph can be created at each granularity. A detailed explanation on the \textbf{scReebTower} construction can be found on Appendix \ref{app:screebtower}. The algorithm is provided in pseudocode below (Algorithm \ref{alg:scReebTower} and \ref{alg:diffusion_condensation}). 

\begin{algorithm}[!h]
\caption{\textsc{scReebTower}}
\label{alg:scReebTower}
\begin{algorithmic}[1]

\Require Dataset \(X\)
\Ensure Reduced multigraph \(\widetilde{S}\)
\State Build the kNN graph on X
\State Create a data kernel based on an adaptive-bandwidth Gaussian kernel as shown in Eq.~\ref{eq:kernel}
\State Compute the data diffusion operator \(P = D^{-1}W\) as in Eq.~\ref{eq:dif_op}
\State Eigendecompose the diffusion operator \(P\) to obtain eigenvectors \(\phi_0, \phi_1, \ldots, \phi_n\)
\State Use the Fiedler vector, $\phi_1$ as a Morse function for Reeb graph construction
\State Construct the Reeb graph by contracting connected components of level sets of $\phi_1(x)$ to their centroids.
\State Reduce the graph to a reduced multigraph \(\widetilde{S}\)

\Return \(\widetilde{S}\)

\end{algorithmic}
\end{algorithm}

\begin{algorithm}[!h]
  \caption{Coarse Graining via Diffusion Condensation}
  \label{alg:diffusion_condensation}
  \begin{algorithmic}[1]
  \Require Data points $X\subset\mathbb{R}^d$, neighbor count $k$, diffusion time $t$
  \Ensure Simplified discrete Reeb graph $G_{\mathrm{simp}}$

  \State Build the adaptive $k_{\mathrm{smooth}}$NN affinity graph on $X$
  \State Compute the diffusion operator $P$ from this affinity graph
  \State Perform the condensation operation $\hat{X}=P^t X$.
  \State Call scReebTower($\hat{X}$)
  \end{algorithmic}
\end{algorithm}

\textbf{scReebTower} has several properties that make it well-suited as a baseline for shape recovery. First, it is topology-agnostic: it does not assume whether the data forms clusters, trees, or cycles. Second, its diffusion-based preprocessing and filter make it robust to noise and non-uniform sampling. Third, the construction is geometry-aware, since the filter is derived from the graph diffusion operator and reflects intrinsic connectivity of the data rather than Euclidean distance. 

\vspace{-.8cm}
\section{Evaluations}
\vspace{-.3cm}
\label{sec:evaluations}
This section defines the evaluation protocols used in \textsc{scShapeBench}. We describe two complementary tracks that mirror the dataset construction: a \emph{synthetic} track, where the generating graph is known and shape recovery can be evaluated directly and a \emph{scRNAseq} track, where methods are  evaluated against expert multi-label shape annotations. Quantitative results for all baselines are reported in Section \ref{sec:results}. 

\vspace{-.3cm}
\subsection{Synthetic Dataset Evaluation}
\vspace{-.3cm}
Given a noisy point cloud \(X_i\), each method outputs a recovered graph \(S_i\). Because the synthetic generator provides the latent graph \(G_i\), evaluation reduces to comparing the reduced graphs through the topological similarity measure $\mathcal{C}(\widetilde{S}_i,\widetilde{G}_i)$ introduced in Section~\ref{sec:problem-formulation}. In practice, we instantiate \(\mathcal{C}\) using two complementary metrics: Wasserstein persistence similarity and graph edit distance.

\subsubsection{Primary metrics}
\vspace{-.3cm}
\noindent\textbf{Wasserstein persistence similarity.}
One instantiation of \(\mathcal{C}(\widetilde{S}_i,\widetilde{G}_i)\) is based on persistent homology under the edge-length filtration described in Section~\ref{sec:background}. We compute persistence diagrams for each reduced graph and compare them using Wasserstein distances~\cite{wassersteinDistances}. Distances are converted into similarity scores so that larger values indicate closer agreement, and we report the mean similarity across all evaluated samples.

\noindent\textbf{Graph edit distance.}
A second instantiation of \(\mathcal{C}(\widetilde{S}_i,\widetilde{G}_i)\) is given by graph edit distance (GED). We compute an approximate GED between reduced multigraphs using the NetworkX~\cite{networkx} implementation of graph edit distance~\cite{grapheditdistance}. Lower values indicate closer combinatorial agreement, and we report the mean GED across all evaluated samples.

These two instantiations of \(\mathcal{C}\) are complementary. Wasserstein persistence similarity measures agreement of multi-scale topological structure, while graph edit distance measures agreement of the reduced graph combinatorics. Strong performance on both therefore indicates recovery of both the global shape and the underlying graph structure.

\subsection{Single-cell RNA sequencing data}
\vspace{-.3cm}
Unlike the synthetic benchmark, real scRNAseq datasets do not admit an exact latent graph \(G\) against which a recovered graph can be directly compared. Consequently, the similarity functional \(\mathcal{C}(\widetilde{S},\widetilde{G})\) introduced in Section~\ref{sec:problem-formulation} cannot be evaluated directly. Instead, expert biologists annotate each dataset with its expected \emph{shape} (clusters, single-trajectory, multi-branching, archetypal) inferred from the source publication and the 2D PHATE/UMAP embeddings described in ~\ref{sec:expert-annotation}. 

We therefore evaluate recovered graphs indirectly through their ability to support prediction of these expert-annotated shape labels. Each method produces a graph \(S\) on the dataset point cloud, and we assess how well $S$ supports recovery of the annotated \textit{shape} under complementary readouts.

\noindent\textbf{Topological readout (PI-MLP).} We compute $H_0$ and $H_1$ persistence barcodes from a normalized edge length filtration on $S$. Barcodes are vectorized as persistence images \citep{adams2017persistence} with a fixed Gaussian kernel bandwidth, weighting function, and pixel resolution shared across all methods, and the resulting feature vectors are passed to a multi-label MLP trained with binary cross-entropy loss. Because the persistence image vectorization is fixed and shared across benchmark entries, any signal recovered is attributable to the geometry of the persistence diagrams themselves, modulated by the representational capacity of the downstream MLP (see Appendix \ref{app:eval_details} for a full description).

\noindent\textbf{Connectivity readout (GNN).} We train a two-layer message-passing GNN network with a graph-level multi-label head on $S$, using a 7-dimensional node featurization (spatial coordinates, degree statistics, clustering coefficient, parallel-edge excess, and component size fraction) and 2-dimensional edge features (normalized weight and edge multiplicity) shared across methods. This readout consumes local connectivity, edge-weight geometry, and attachment patterns, and tests whether the recovered graph encodes the annotated shape in a form that a small learned model can decode.

\vspace{-.4cm}
\section{Empirical Results}
\vspace{-.3cm}
\label{sec:results}
We report results on the two tracks defined in Section \ref{sec:datasets}. The synthetic track measures shape recovery against known ground-truth graphs; the scRNAseq track measures recovery of expert shape annotations through the classifier readouts defined in Section \ref{sec:evaluations}. Visualization examples are provided in Appendix~\ref{app:visu_graphs}.

\subsection{Synthetic Results}
\vspace{-.3cm}
Table~\ref{tab:benchmark-latent-transposed} reports performance on the synthetic track using the two primary graph-similarity metrics in our final evaluation. \textbf{scReebTower} achieves the best Wasserstein persistence similarity and the lowest graph edit distance, slightly improving over the base \texttt{scReebTower ($\ell$=0)} on both metrics. \texttt{PAGA} remains competitive in persistence similarity but has higher graph edit distance, while the Mapper-family methods (\texttt{mapper}) perform substantially worse on both measures.

\begin{table}[H]
  \centering
  \caption{Benchmark results on the synthetic track. Higher is better for Wasserstein persistence similarity, lower is better for graph edit distance}
  \label{tab:benchmark-latent-transposed}
  \setlength{\tabcolsep}{10pt}
  \scriptsize
  \resizebox{\textwidth}{!}{%
  \begin{tabular}{lccccc}
  \toprule
  Metric & \texttt{scReebTower ($\ell$=0)} 
  &
  \texttt{scReebTower} 
  &
  \texttt{paga} & \texttt{mapper} & \\
  \midrule
  Wasserstein Persistence Similarity & 0.2806 & \textbf{0.3199} & 0.2788 & 0.0318 \\
  Graph Edit Distance & 13.3378 & \textbf{13.3216} & 16.2518 & 79.7820 
  \\
  \bottomrule
  \end{tabular}%
  }
\end{table}

\subsection{scRNAseq Results}

For the scRNAseq data, performance variance across methods highlights how biological annotation complexity can confound evaluation, directly motivating our inclusion of synthetic benchmarks for ground-truth validation. As reported in Table~\ref{tab:scrnaseq-results}, the scRNAseq track serves as a rigorous baseline. \textbf{scReebTower} achieves the best accuracy through the different shape classes. These results validate the benchmark's ability to resolve performance differences even under varying architectural constraints.


\begin{table}[H]
\centering
\caption{Per-shape accuracy and F1 on the scRNAseq track of \textsc{scShapeBench}. Shapes: \textbf{A}rchetypal, \textbf{Cl}usters, \textbf{S}imple \textbf{T}rajectory, \textbf{M}ultiple \textbf{T}rajectories. Best per column in \textbf{bold}; second best \underline{underlined}. Values are mean $\pm$ std over 5 CV folds.}
\label{tab:scrnaseq-results}
\setlength{\tabcolsep}{3pt}
\resizebox{\textwidth}{!}{%
\begin{tabular}{l cccccccc c cccccccc}
\toprule
& \multicolumn{8}{c}{GNN} &  & \multicolumn{8}{c}{MLP} \\
\cmidrule(lr){2-9} \cmidrule(lr){11-18}
 & \multicolumn{2}{c}{Cl} & \multicolumn{2}{c}{ST} & \multicolumn{2}{c}{MT} & \multicolumn{2}{c}{A} &  & \multicolumn{2}{c}{Cl} & \multicolumn{2}{c}{ST} & \multicolumn{2}{c}{MT} & \multicolumn{2}{c}{A} \\
\cmidrule(lr){2-3} \cmidrule(lr){4-5} \cmidrule(lr){6-7} \cmidrule(lr){8-9} \cmidrule(lr){11-12} \cmidrule(lr){13-14} \cmidrule(lr){15-16} \cmidrule(lr){17-18}
Method & Acc & F1 & Acc & F1 & Acc & F1 & Acc & F1 &  & Acc & F1 & Acc & F1 & Acc & F1 & Acc & F1 \\
\midrule
mapper & .61$\pm$.08 & .71$\pm$.07 & .57$\pm$.06 & .70$\pm$.06 & .76$\pm$.05 & .86$\pm$.04 & .71$\pm$.14 & .78$\pm$.12 &  & \underline{.65$\pm$.08} & \underline{.78$\pm$.07} & .64$\pm$.09 & .75$\pm$.08 & .92$\pm$.07 & .96$\pm$.04 & .73$\pm$.11 & .84$\pm$.08 \\
paga & \underline{.69$\pm$.01} & \underline{.78$\pm$.01} & .70$\pm$.06 & \textbf{.81$\pm$.05} & \underline{.80$\pm$.08} & \underline{.88$\pm$.05} & \underline{.73$\pm$.08} & \textbf{.81$\pm$.06} &  & .65$\pm$.10 & .76$\pm$.09 & \underline{.67$\pm$.14} & \underline{.77$\pm$.11} & .91$\pm$.09 & .95$\pm$.05 & \textbf{.74$\pm$.10} & \underline{.84$\pm$.07} \\
scReeb & .57$\pm$.06 & .71$\pm$.04 & \textbf{.70$\pm$.10} & \underline{.80$\pm$.08} & \textbf{.83$\pm$.13} & \textbf{.90$\pm$.08} & .62$\pm$.07 & .74$\pm$.05 &  & \textbf{.68$\pm$.05} & \textbf{.81$\pm$.04} & \textbf{.69$\pm$.11} & \textbf{.79$\pm$.09} & \underline{.95$\pm$.04} & \underline{.97$\pm$.02} & \underline{.74$\pm$.11} & \textbf{.84$\pm$.08} \\
scReebTower & \textbf{.75$\pm$.10} & \textbf{.82$\pm$.09} & \underline{.70$\pm$.13} & .77$\pm$.10 & .21$\pm$.29 & .26$\pm$.34 & \textbf{.77$\pm$.08} & \underline{.80$\pm$.08} &  & .56$\pm$.13 & .70$\pm$.12 & .61$\pm$.18 & .69$\pm$.17 & \textbf{.96$\pm$.05} & \textbf{.98$\pm$.02} & .56$\pm$.13 & .67$\pm$.16 \\
\bottomrule
\end{tabular}%
}
\end{table}

\vspace{-.5cm}
\section{Conclusion}
\vspace{-.4cm}
In this paper we introduced \textsc{scShapeBench}, a benchmark for the problem of single-cell shape detection: recovering a reduced graph that reflects the data organization from a high-dimensional point cloud. This problem sits upstream of nearly every scRNAseq analysis pipeline, yet has historically been resolved through manual visual inspection by bioinformaticians, a step that becomes a bottleneck as analysis is increasingly automated and as agentic AI scientists are deployed on biological data.

\textsc{scShapeBench} provides a formal mathematical framework for shape detection, synthetic and expert-annotated datasets, topology-aware metrics, and the scReebTower baseline. Together, these establish a standardized point-cloud-in, topology-out evaluation protocol and demonstrate that the benchmark is not only diagnostic but also enables the development of better methods. The benchmark is also designed to scale: through our dedicated annotation web interface, the expert-annotated corpus can grow over time, incorporating new scRNA-seq datasets to ensure long-term coverage and diversity. We hope \textsc{scShapeBench} provides both a foundation for shape-detection research and a common ground on which new methods can be developed and compared. A detailed discussion of limitations and broader impact is provided in Appendix~\ref{app:limitations}.

\bibliographystyle{unsrt}
\bibliography{references}

@article{Phate,
  author    = {Kevin R. Moon and David van Dijk and Zheng Wang and Scott Gigante and
               Daniel B. Burkhardt and William S. Chen and Kristina Yim and
               Antonia van den Elzen and Matthew J. Hirn and Ronald R. Coifman and
               Natalia B. Ivanova and Guy Wolf and Smita Krishnaswamy},
  title     = {Visualizing structure and transitions in high-dimensional biological data},
  journal   = {Nature Biotechnology},
  year      = {2019},
  volume    = {37},
  number    = {12},
  pages     = {1482--1492},
  doi       = {10.1038/s41587-019-0336-3},
  url       = {https://doi.org/10.1038/s41587-019-0336-3},
}

@article{Umap,
  author    = {Leland McInnes and John Healy and James Melville},
  title     = {UMAP: Uniform Manifold Approximation and Projection for Dimension Reduction},
  journal   = {arXiv preprint arXiv:1802.03426},
  year      = {2018},
  url       = {https://arxiv.org/abs/1802.03426},
}

@inproceedings{MapperPaper,
booktitle = {Eurographics Symposium on Point-Based Graphics},
editor = {M. Botsch and R. Pajarola and B. Chen and M. Zwicker},
title = {{Topological Methods for the Analysis of High Dimensional Data Sets and 3D Object Recognition}},
author = {Singh, Gurjeet and Memoli, Facundo and Carlsson, Gunnar},
year = {2007},
publisher = {The Eurographics Association},
ISSN = {1811-7813},
ISBN = {978-3-905673-51-7},
DOI = {/10.2312/SPBG/SPBG07/091-100}
}

@misc{smitaDiffusion,
      title={Time-inhomogeneous diffusion geometry and topology}, 
      author={Guillaume Huguet and Alexander Tong and Bastian Rieck and Jessie Huang and Manik Kuchroo and Matthew Hirn and Guy Wolf and Smita Krishnaswamy},
      year={2023},
      eprint={2203.14860},
      archivePrefix={arXiv},
      primaryClass={cs.LG},
      url={https://arxiv.org/abs/2203.14860}, 
}

@inproceedings{smitaDiffusion2,
   title={Coarse Graining of Data via Inhomogeneous Diffusion Condensation},
   url={http://dx.doi.org/10.1109/BigData47090.2019.9006013},
   DOI={10.1109/bigdata47090.2019.9006013},
   booktitle={2019 IEEE International Conference on Big Data (Big Data)},
   publisher={IEEE},
   author={Brugnone, Nathan and Gonopolskiy, Alex and Moyle, Mark W. and Kuchroo, Manik and Dijk, David van and Moon, Kevin R. and Colon-Ramos, Daniel and Wolf, Guy and Hirn, Matthew J. and Krishnaswamy, Smita},
   year={2019},
   month=dec, pages={2624–2633} }

@Article{Haghverdi2016,
author={Haghverdi, Laleh
and B{\"u}ttner, Maren
and Wolf, F. Alexander
and Buettner, Florian
and Theis, Fabian J.},
title={Diffusion pseudotime robustly reconstructs lineage branching},
journal={Nature Methods},
year={2016},
month={Oct},
day={01},
volume={13},
number={10},
pages={845-848},
issn={1548-7105},
doi={10.1038/nmeth.3971},
url={https://doi.org/10.1038/nmeth.3971}
}

@article{Saelens2019,
  title = {A comparison of single-cell trajectory inference methods},
  volume = {37},
  ISSN = {1546-1696},
  url = {http://dx.doi.org/10.1038/s41587-019-0071-9},
  DOI = {10.1038/s41587-019-0071-9},
  number = {5},
  journal = {Nature Biotechnology},
  publisher = {Springer Science and Business Media LLC},
  author = {Saelens,  Wouter and Cannoodt,  Robrecht and Todorov,  Helena and Saeys,  Yvan},
  year = {2019},
  month = apr,
  pages = {547–554}
}

@article{czi2025cz,
  title={CZ CELLxGENE Discover: a single-cell data platform for scalable exploration, analysis and modeling of aggregated data},
  author={CZI Cell Science Program and Abdulla, Shibla and Aevermann, Brian and Assis, Pedro and Badajoz, Seve and Bell, Sidney M and Bezzi, Emanuele and Cakir, Batuhan and Chaffer, Jim and Chambers, Signe and others},
  journal={Nucleic acids research},
  volume={53},
  number={D1},
  pages={D886--D900},
  year={2025},
  publisher={Oxford University Press}
}

@ARTICLE{monocle3,
  title    = "The single-cell transcriptional landscape of mammalian
              organogenesis",
  author   = "Cao, Junyue and Spielmann, Malte and Qiu, Xiaojie and Huang,
              Xingfan and Ibrahim, Daniel M and Hill, Andrew J and Zhang, Fan
              and Mundlos, Stefan and Christiansen, Lena and Steemers, Frank J
              and Trapnell, Cole and Shendure, Jay",
  journal  = "Nature",
  volume   =  566,
  number   =  7745,
  pages    = "496--502",
  month    =  feb,
  year     =  2019,
  address  = "England",
  language = "en"
}

@misc{mioflow,
  doi = {10.48550/ARXIV.2206.14928},
  url = {https://arxiv.org/abs/2206.14928},
  author = {Huguet,  Guillaume and Magruder,  D. S. and Tong,  Alexander and Fasina,  Oluwadamilola and Kuchroo,  Manik and Wolf,  Guy and Krishnaswamy,  Smita},
  title = {Manifold Interpolating Optimal-Transport Flows for Trajectory Inference},
  publisher = {arXiv},
  year = {2022},
  copyright = {arXiv.org perpetual,  non-exclusive license}
}

@ARTICLE{seurat,
  title    = "Spatial reconstruction of single-cell gene expression data",
  author   = "Satija, Rahul and Farrell, Jeffrey A and Gennert, David and
              Schier, Alexander F and Regev, Aviv",
  journal  = "Nat Biotechnol",
  volume   =  33,
  number   =  5,
  pages    = "495--502",
  month    =  apr,
  year     =  2015,
  address  = "United States",
  language = "en"
}

@ARTICLE{PAGA,
  title    = "{PAGA}: graph abstraction reconciles clustering with trajectory
              inference through a topology preserving map of single cells",
  author   = "Wolf, F Alexander and Hamey, Fiona K and Plass, Mireya and
              Solana, Jordi and Dahlin, Joakim S and G{\"o}ttgens, Berthold and
              Rajewsky, Nikolaus and Simon, Lukas and Theis, Fabian J",
  journal  = "Genome Biol",
  volume   =  20,
  number   =  1,
  pages    = "59",
  month    =  mar,
  year     =  2019,
  address  = "England",
  language = "en"
}

@article{wassersteinDistances,
doi = {10.1088/0266-5611/27/12/124007},
url = {https://doi.org/10.1088/0266-5611/27/12/124007},
year = {2011},
month = {nov},
publisher = {},
volume = {27},
number = {12},
pages = {124007},
author = {Mileyko, Yuriy and Mukherjee, Sayan and Harer, John},
title = {Probability measures on the space of persistence diagrams},
journal = {Inverse Problems}
}

@inproceedings{grapheditdistance,
  TITLE = {{An Exact Graph Edit Distance Algorithm for Solving Pattern Recognition Problems}},
  AUTHOR = {Abu-Aisheh, Zeina and Raveaux, Romain and Ramel, Jean-Yves and Martineau, Patrick},
  URL = {https://hal.science/hal-01168816},
  BOOKTITLE = {{4th International Conference on Pattern Recognition Applications and Methods 2015}},
  ADDRESS = {Lisbon, Portugal},
  YEAR = {2015},
  MONTH = Jan,
  DOI = {10.5220/0005209202710278},
  HAL_ID = {hal-01168816},
  HAL_VERSION = {v1},
}

@inproceedings{networkx,
author = {Hagberg, Aric and Swart, Pieter and Chult, Daniel},
year = {2008},
month = {06},
pages = {},
title = {Exploring Network Structure, Dynamics, and Function Using NetworkX},
journal = {Proceedings of the 7th Python in Science Conference},
doi = {10.25080/TCWV9851}
}

@article{Reeb1946,
  author  = {Reeb, Georges},
  title   = {Sur les points singuliers d'une forme de Pfaff compl{\`e}tement int{\'e}grable ou d'une fonction num{\'e}rique},
  journal = {Comptes Rendus de l'Acad{\'e}mie des Sciences},
  volume  = {222},
  pages   = {847--849},
  year    = {1946}
}

@article{tarhan2023single,
  title={Single Cell Portal: an interactive home for single-cell genomics data},
  author={Tarhan, Leyla and Bistline, Jon and Chang, Jean and Galloway, Bryan and Hanna, Emily and Weitz, Eric},
  journal={BioRxiv},
  year={2023}
}

@article{george2024expression,
  title={Expression Atlas update: insights from sequencing data at both bulk and single cell level},
  author={George, Nancy and Fexova, Silvie and Fuentes, Alfonso Munoz and Madrigal, Pedro and Bi, Yalan and Iqbal, Haider and Kumbham, Upendra and Nolte, Nadja Francesca and Zhao, Lingyun and Thanki, Anil S and others},
  journal={Nucleic Acids Research},
  volume={52},
  number={D1},
  pages={D107--D114},
  year={2024},
  publisher={Oxford University Press}
}

@article{wolf2018scanpy,
  title={SCANPY: large-scale single-cell gene expression data analysis},
  author={Wolf, F Alexander and Angerer, Philipp and Theis, Fabian J},
  journal={Genome biology},
  volume={19},
  number={1},
  pages={15},
  year={2018},
  publisher={Springer}
}

@article{kuchroo2022multiscale,
  title={Multiscale PHATE identifies multimodal signatures of COVID-19},
  author={Kuchroo, Manik and Huang, Jessie and Wong, Patrick and Grenier, Jean-Christophe and Shung, Dennis and Tong, Alexander and Lucas, Carolina and Klein, Jon and Burkhardt, Daniel B and Gigante, Scott and others},
  journal={Nature biotechnology},
  volume={40},
  number={5},
  pages={681--691},
  year={2022},
  publisher={Nature Publishing Group US New York}
}

@article{krishnaswamy2023revealing,
  title={Revealing dynamic temporal regulatory networks driving cancer cell state plasticity with neural ODE-based optimal transport},
  author={Krishnaswamy, Smita and Tong, Alex and Kuchroo, Manik and Gupta, Shabarni and Venkat, Aarthi and San Juan, Beatriz and Rangel, Laura and Zhu, Brandon and Lock, John and Chaffer, Christine},
  year={2023}
}

@article{milner2020heterogenous,
  title={Heterogenous populations of tissue-resident CD8+ T cells are generated in response to infection and malignancy},
  author={Milner, J Justin and Toma, Clara and He, Zhaoren and Kurd, Nadia S and Nguyen, Quynh P and McDonald, Bryan and Quezada, Lauren and Widjaja, Christella E and Witherden, Deborah A and Crowl, John T and others},
  journal={Immunity},
  volume={52},
  number={5},
  pages={808--824},
  year={2020},
  publisher={Elsevier}
}

@article{coifman2006diffusion,
  title={Diffusion maps},
  author={Coifman, Ronald R and Lafon, St{\'e}phane},
  journal={Applied and computational harmonic analysis},
  volume={21},
  number={1},
  pages={5--30},
  year={2006},
  publisher={Elsevier}
}

@article{venkat2025aanet,
  title={AAnet Resolves a Continuum of Spatially Localized Cell States to Unveil Intratumoral Heterogeneity},
  author={Venkat, Aarthi and Youlten, Scott E and San Juan, Beatriz P and Purcell, Carley A and Gupta, Shabarni and Amodio, Matthew and Neumann, Daniel P and Lock, John G and Westacott, Anton E and McCool, Cerys S and others},
  journal={Cancer Discovery},
  volume={15},
  number={10},
  pages={2139--2165},
  year={2025},
  publisher={American Association for Cancer Research}
}

@article{adams2017persistence,
  title={Persistence images: A stable vector representation of persistent homology},
  author={Adams, Henry and Emerson, Tegan and Kirby, Michael and Neville, Rachel and Peterson, Chris and Shipman, Patrick and Chepushtanova, Sofya and Hanson, Eric and Motta, Francis and Ziegelmeier, Lori},
  journal={Journal of Machine Learning Research},
  volume={18},
  number={8},
  pages={1--35},
  year={2017}
}

@article{xu2018powerful,
  title={How powerful are graph neural networks?},
  author={Xu, Keyulu and Hu, Weihua and Leskovec, Jure and Jegelka, Stefanie},
  journal={arXiv preprint arXiv:1810.00826},
  year={2018}
}

@article{hu2019strategies,
  title={Strategies for pre-training graph neural networks},
  author={Hu, Weihua and Liu, Bowen and Gomes, Joseph and Zitnik, Marinka and Liang, Percy and Pande, Vijay and Leskovec, Jure},
  journal={arXiv preprint arXiv:1905.12265},
  year={2019}
}

@inproceedings{tong2020trajectorynet,
  title={Trajectorynet: A dynamic optimal transport network for modeling cellular dynamics},
  author={Tong, Alexander and Huang, Jessie and Wolf, Guy and Van Dijk, David and Krishnaswamy, Smita},
  booktitle={International conference on machine learning},
  pages={9526--9536},
  year={2020},
  organization={PMLR}
}

@article{anchang2016visualization,
  title={Visualization and cellular hierarchy inference of single-cell data using SPADE},
  author={Anchang, Benedict and Hart, Tom DP and Bendall, Sean C and Qiu, Peng and Bjornson, Zach and Linderman, Michael and Nolan, Garry P and Plevritis, Sylvia K},
  journal={Nature protocols},
  volume={11},
  number={7},
  pages={1264--1279},
  year={2016},
  publisher={Nature Publishing Group UK London}
}

@inproceedings{persistentstab,
author = {Cohen-Steiner, David and Edelsbrunner, Herbert and Harer, John},
year = {2005},
month = {06},
pages = {263-271},
title = {Stability of Persistence Diagrams},
volume = {37},
journal = {Discrete & Computational Geometry - DCG},
doi = {10.1007/s00454-006-1276-5}
}

@ARTICLE{Hubert1985-tk,
  title    = "Comparing partitions",
  author   = "Hubert, Lawrence and Arabie, Phipps",
  journal  = "Journal of Classification",
  volume   =  2,
  number   =  1,
  pages    = "193--218",
  month    =  dec,
  year     =  1985
}

\newpage
\appendix

\section{Related Work}
\label{app:related_works}
\noindent\textbf{Graph-based and topological structure recovery.}
Many methods for recovering structure from high-dimensional data begin by constructing a neighborhood graph, most commonly a \(k\)-nearest-neighbor graph. Graph abstraction methods such as PAGA \citep{PAGA} summarize connectivity between cell populations, while trajectory inference methods such as SPADE \citep{anchang2016visualization} and Monocle \citep{monocle3} construct developmental paths from local relationships between cells. In contrast, topological approaches such as Mapper \citep{MapperPaper} build graph summaries from level sets of a filter function rather than from a global neighborhood graph. These approaches reflect different paradigms for recovering intrinsic structure from point clouds.

\noindent\textbf{Evaluation and benchmarks.}
Evaluation protocols for structure recovery remain fragmented and task-specific. Clustering methods are typically evaluated using partition metrics \citep{Hubert1985-tk}, which ignore global connectivity, while trajectory inference benchmarks such as dynverse \citep{Saelens2019} focus on pseudotime and lineage recovery under tree-like assumptions. Graph abstraction methods are often assessed qualitatively, and topological data analysis methods emphasize stability guarantees \citep{persistentstab} rather than structural reconstruction accuracy. Existing benchmarks therefore do not evaluate recovery across multiple topology classes using metrics that capture both topological fidelity and downstream utility.

\noindent\textbf{Diffusion-based representations.}
Diffusion-based methods provide geometry-aware representations of high-dimensional data by modeling connectivity over multiple scales. Diffusion maps \citep{coifman2006diffusion} construct coordinates from eigenvectors of a diffusion operator that approximates the Laplace--Beltrami operator of the underlying manifold. These representations are robust to noise and have been widely used in single-cell analysis, including diffusion pseudotime \citep{Haghverdi2016} and PHATE \citep{phate}, for recovering continuous and branching biological structure.

Taken together, prior work lacks a unified framework for evaluating general intrinsic structure recovery from high-dimensional point clouds, motivating the development of \textsc{scShapeBench}.

\section{scRNAseq corpus}
\subsection{Corpus composition and per-dataset metadata}
\label{app:corpus}
The corpus comprises 102 datasets drawn from four sources: CELLxGENE Census (62), 10x Genomics reference catalog (30), the Broad Single Cell Portal (8), and the EMBL-EBI Single Cell Expression Atlas (1).

The corpus covers $7$~organisms (predominantly \textit{Homo sapiens} and \textit{Mus musculus}, with rat, rhesus macaque, zebrafish, and mixed human/mouse cell-line samples), $39$~tissues, and a range of assays including 10x Chromium 3$'$ v2/v3, 10x 5$'$ v1/v2, 10x~GEM-X, 10x~NextGEM, and 10x~Flex/scFFPE. Per-dataset cell counts range from $1{,}163$ to $83{,}943$, with a median of $20{,}556$. We provide a spreadsheet of the dataset details in the supplementary material containing recording dataset ID, publication link, species, tissue, cell type focus, disease condition, sequencing technology, cell and gene counts, DOI, raw data format and are hosted on Hugging Face.

\subsection{Preprocessing}
\label{app:preprocessing}

For each dataset we (1)~load the raw count matrix from the source \texttt{h5ad} or convert it from the source format; (2)~drop cells with fewer than $100$ detected genes and drop genes detected in fewer than $100$ cells; (3)~optionally apply a mitochondrial-fraction filter when mitochondrial annotations are available (off by default; reported per dataset); (4)~normalize each cell to a target count of $10{,}000$ and apply $\log(x+1)$ transformation; (5)~when the dataset exceeds $50{,}000$ cells, uniformly subsample to $50{,}000$ with random seed $42$, optionally stratifying by an \texttt{obs} label column when one is available; and (6)~enrich the \texttt{.obs} table with a unique \texttt{dataset\_id}, per-cell gene count, and per-dataset cell count. Highly-variable-gene selection is implemented (Seurat~v3 flavor) but disabled by default in the released corpus, so that downstream methods can choose their own feature-selection strategy. The full configuration is exposed as a \texttt{PreprocessingConfig} dataclass and the entry point is a single command, \texttt{python scripts/preprocess.py}, which writes a per-dataset report alongside each output file. The random seed ($42$) is the only stochastic input to the pipeline and is shared across the subsampling step and any downstream embedding computed from the preprocessed file.

\subsection{Annotation Aggregation}
\label{app:aggregation}

Because datasets may exhibit multiple organizational regimes simultaneously, we evaluate agreement at both the label-set and per-label levels. We summarize label-set agreement using the average pairwise Jaccard similarity between annotators, and summarize per-label agreement by treating each label as an independent present/absent decision (Table~\ref{tab:annotation-agreement}). Across the corpus, the mean pairwise Jaccard similarity was \(0.403\), while per-label Fleiss' \(\kappa\) ranged from \(0.161\) to \(0.255\).

These agreement levels reflect the inherent ambiguity of real scRNA-seq datasets, where multiple reasonable downstream analyses may apply simultaneously. In practice, biologists frequently choose analysis pipelines based on subjective interpretation of embeddings and biological context, and disagreement between annotators often corresponds to legitimate uncertainty about which downstream methods are most appropriate rather than annotation error.

For this reason, \textsc{scShapeBench} uses union aggregation for the primary evaluation: a shape label is considered positive if it is selected by at least one annotator. This aggregation strategy reflects the benchmark objective of recovering the set of biologically plausible structural interpretations of a dataset rather than enforcing a single canonical topology.

We additionally release the full multi-annotator label distributions, agreement statistics, and alternative aggregation rules alongside the benchmark datasheet. The 2D embeddings provided to annotators serve strictly as visual aids for human interpretation and are excluded from all benchmarked recovery methods and downstream evaluations (Section~\ref{sec:results}).

\begin{table}[h!]
  \centering
  \caption{Unweighted inter-annotator agreement for the simplified scRNA-seq shape taxonomy.
  Prevalence is the fraction of datasets for which at least one annotator selected the label.
  Mean support is the average number of annotators, out of nine, selecting the label per dataset.
  Agreement and Fleiss' \(\kappa\) are computed by treating each label as a binary
  present/absent decision.}
  \label{tab:annotation-agreement}
  \begin{tabular}{lrrrr}
  \toprule
  Label & Prevalence & Mean support & Agreement & Fleiss' \(\kappa\) \\
  \midrule
  Clusters & 0.716 & 2.147 & 0.722 & 0.234 \\
  Single trajectory & 0.735 & 2.373 & 0.711 & 0.255 \\
  Multi-branching & 0.912 & 3.971 & 0.614 & 0.218 \\
  Archetypal & 0.755 & 2.235 & 0.687 & 0.161 \\
  \bottomrule
 \end{tabular}
\end{table}

\section{scReebTower algorithm}

\noindent\textbf{Construction.} 

Let \(G_{\mathrm{kNN}} = (V,E)\) denote the \(k\)-nearest-neighbor graph and let
\[
f : V \to \mathbb{R}
\]
be the discrete Morse function induced by the leading non-trivial diffusion map.

Order the vertex values
\[
f(v_1) \leq f(v_2) \leq \cdots \leq f(v_n),
\]
and define threshold values
\[
a_i = \frac{f(v_i) + f(v_{i+1})}{2}.
\]

For each threshold \(a_i\), the level set \(f^{-1}(a_i)\) is defined by the subgraph consisting of edges \((u,v) \in E\) satisfying
\[
f(u) \leq a_i < f(v)
\quad\text{or}\quad
f(v) \leq a_i < f(u).
\]
This process is visualized in  \ref{fig:reebalg}.

Connected components of this level-set subgraph define nodes of the Reeb graph. Nodes corresponding to adjacent thresholds \(a_i\) and \(a_{i+1}\) are connected whenever their underlying level-set components share points.

Because connectivity of the level sets can change only when crossing a vertex value of \(f\), examining one threshold between consecutive vertex values suffices to recover the Reeb graph.

The resulting graph tracks the evolution of connected components across the Morse function and produces a graph representation of the underlying data shape. Finally, maximal chains of degree-two vertices are contracted to produce the reduced Reeb graph.
\paragraph{Time complexity.}
  For the full scReebTower pipeline, the dominant step is computing the
  diffusion
  eigenfunction. In the worst case, this requires dense eigendecomposition of an
  \(n \times n\) diffusion operator, giving overall time complexity
  \[
  O(n^3).
  \]
  After the filter values are computed, the Reeb graph construction itself runs
  in
  \[
  O(E \log S + Sn),
  \]
  where \(E\) is the number of edges in the neighborhood graph and \(S\leq n-1\)
  is the number of midpoint slices induced by the unique filter values. For a
  \(k\)-nearest-neighbor graph, \(E=O(kn)\), so this becomes
  \[
  O(kn\log n+n^2).
  \]
  Thus the full worst-case runtime is dominated by the eigendecomposition term,
  \(O(n^3)\).

\label{app:screebtower}

\section{Evaluation Implementation Details}
\label{app:eval_details}

All real-data readouts are evaluated with stratified 5-fold cross-validation, with folds fixed across methods for comparability. Model random seeds are set per fold. Per-class performance is reported as AUC, AUPRC, and F$_1$; macro-averages over the four shape classes are used as summary statistics.

\subsection*{Persistence image featurization}

$H_0$ and $H_1$ barcodes are extracted from the edge-weight filtration on $S$, with edge weights normalized to $[0,1]$ by the graph's maximum weight. $H_0$ bars are encoded as birth--persistence pairs $(0,\, d_i)$ where $d_i$ is the finite death value; $H_1$ bars are essential and encoded as $(b_i,\, 1-b_i)$. Each barcode is vectorized as a persistence image \citep{adams2017persistence} on a $10\times10$ grid over $[0,1]^2$ with Gaussian kernel bandwidth $\sigma=0.1$ and linear persistence weighting $w(b,p)=p$. The flattened $H_0$ and $H_1$ images are concatenated to form a 200-dimensional vector. This featurization is identical for PI-MLP and PI-SVM.

\paragraph{PI-MLP.} A single-hidden-layer MLP (64 units, ReLU) operating on 200-dimensional persistence image features (described below). Features are $z$-score normalized using training-fold statistics. Trained with binary cross-entropy, Adam (learning rate $10^{-3}$, weight decay $10^{-4}$), batch size 16, for 1 epoch.

\paragraph{GNN.} A two-layer GINEConv \citep{xu2018powerful,hu2019strategies} network with hidden dimension 32 and dropout $p=0.3$. Node features are 7-dimensional: 2D spatial coordinates (first two columns of the point cloud, $z$-score normalized per graph), raw normalized degree, deduplicated normalized degree, incident parallel-edge excess, local clustering coefficient, and fractional connected-component size. Edge features are 2-dimensional: normalized edge weight and normalized parallel-edge multiplicity. After each convolution, batch normalization and ReLU are applied. Graph-level representations are formed by concatenating global mean and max pooling over node embeddings, then projected through a two-layer MLP head to four binary outputs. Training uses binary cross-entropy with per-class positive-frequency reweighting, Adam (learning rate $10^{-3}$, weight decay $10^{-4}$), gradient clipping (max norm 1.0), and 100 epochs with batch size 16.

\subsection{Synthetic Data Relevant Hyperparameters}
\label{app:hyperparameters}

All methods were run on the input point cloud using fixed hyperparameters across benchmark samples, with no per-sample tuning.

\paragraph{scReebTower (base).}
The base \textsc{scReebTower} configuration used a \(k\)-nearest-neighbor graph with \(k=15\), the diffusion eigenfunction as the filter function. Edge-length output was enabled for persistence-based evaluation.

\paragraph{scReebTower (multiscale).}
The multiscale \textsc{scReebTower} configuration used the same Reeb graph construction parameters together with diffusion condensation prior to graph construction. Reeb graph construction used \(k=15\), while condensation smoothing used
\[
k_{\mathrm{smooth}}=\min(80,n-1),
\]
where \(n\) is the number of input points. The diffusion eigenfunction was used as the filter function, Edge-length output was retained for persistence evaluation.

\paragraph{PAGA.}
PAGA was run using Scanpy with Leiden clustering followed by PAGA connectivity estimation. Neighborhood graphs used

$k_{\mathrm{PAGA}}=\min(\max(15,2),n-1),$
with the input point cloud \(X\) used as the neighbor representation. Leiden clustering used resolution \(1.0\), random seed \(0\), the \texttt{igraph} backend, two optimization iterations, and an undirected graph. PAGA groups were defined by Leiden clusters, and all edges with connectivity greater than zero were retained.

\paragraph{Mapper.}
Mapper was run using KeplerMapper with a PCA lens and DBSCAN clustering inside cover elements. The PCA lens dimension was

$d_{\mathrm{lens}}=\min(2,d,n),$
where \(d\) is the ambient dimension and \(n\) is the number of input points. PCA used random seed \(0\). The cover consisted of six cubical intervals per lens dimension with overlap fraction \(0.35\). DBSCAN clustering used minimum samples \(3\) and radius

$\epsilon = 1.5 \cdot \operatorname{median}_{i}\left(r_i^{(3)}\right),$
where \(r_i^{(3)}\) denotes the distance from point \(i\) to its third-nearest neighbor. Duplicate Mapper nodes were removed.

\subsection{Synethic Benchmark Algorithm Summary}
  \begin{algorithm}[H]
  \caption{Synthetic dataset generation with controlled topology and difficulty}
  \label{alg:graph_generation}
  \begin{algorithmic}[1]
  \Require Number of samples $N$, difficulty preset, seed
  \Ensure Noisy point clouds \(X_i\), target graphs \(G_i\), and metadata

  \For{$i = 1$ to $N$}
      \Repeat
          \State Sample difficulty coordinates: requested noise, crowding,
  density, and thickness budget
          \State Sample requested embedding dimension \(d \in \{2,3,4,5\}\)
          \State Sample the number of connected components
          \For{each component}
              \State Sample a topology class
              \State Generate a cycle backbone, if required by the topology
  class
              \State Attach acyclic branches subject to degree and size caps
              \State Constructively embed the component in \(\mathbb{R}^d\)
          \EndFor
          \State Assemble components into a disconnected graph with controlled
  gaps
          \State Choose an admissible tube radius from the thickness budget
          \State Apply thickness-aware PCA crowding
          \State Cap the Gaussian noise scale by the post-crowding clearance
          \State Sample points from solid edge tubes and junction nodes
          \State Add Gaussian noise to obtain \(X_i\)
          \State Validate separation, tube-overlap, component-count, and noise-
  confusion constraints
      \Until{all constraints are satisfied}
      \State Save \(X_i\), \(G_i\), embedded positions, requested coordinates,
  realized geometry, and topology metadata
  \EndFor
  \end{algorithmic}
  \end{algorithm}

\paragraph{Synthetic benchmark generator.}
The synthetic benchmark was generated using the full difficulty preset with random seed \(20260422\). The benchmark contained \(1000\) samples with at most \(100\) rejected generation attempts per sample. Feature scale was fixed at \(1.0\), endpoint sampling mode was set to \texttt{include}, node samples were retained, and at least two points were sampled per edge. Density jitter was sampled in \([0.0,0.15]\), and the minimum structural separation parameter was fixed at
\[
\epsilon_{\mathrm{sep}}=0.02.
\]

The full difficulty preset sampled requested noise ratios in \([0.02,0.20]\), separation parameters in \([0.00,0.60]\), sampling densities in \([3.5,25.0]\), and feature-thickness budgets in \([0.45,1.00]\).

Topology sampling used the graph classes \texttt{singleton}, \texttt{single\_edge}, \texttt{tree}, \texttt{single\_cycle}, \texttt{multi\_cycle}, and \texttt{hybrid} with probabilities
\[
(0.03,\,0.05,\,0.17,\,0.20,\,0.25,\,0.30).
\]
Cycle lengths ranged from \(3\) to \(8\), with at most six cycles per component. Branch attachment counts ranged from \(0\) to \(4\), branch lengths ranged from \(1\) to \(4\), maximum branch depth ranged from \(1\) to \(3\), and graph degree was capped at \(5\). Samples contained at most \(80\) nodes and \(88\) edges.

Disconnected examples were generated by recursively adding connected components with initial probability \(0.25\), geometric decay factor \(0.55\), and maximum of four components. Component gap ratios ranged from \(4.0\) to \(8.0\), with at least \(20\) points per component.

Embedding dimensions were sampled from \(\{2,3,4,5\}\) with probabilities
\[
(0.50,\,0.35,\,0.10,\,0.05).
\]
The base embedding separation-to-feature ratio was \(6.0\), edge-length jitter ranged from \(0.0\) to \(0.12\), and branch-angle spread ranged from \(0.5\) to \(1.2\). Tube sampling used solid tubes with minimum realized edge-tube radius ratio \(0.02\) and cycle-hole radius cap fraction \(0.20\).

Noise-clearance and separation enforcement used isotropy epsilon \(0.05\), minimum separation gamma \(0.01\), search tolerance \(0.005\), maximum search iterations \(24\), and safety multipliers \(6.0\) for both noise clearance and disconnected component separation.

\section{Limitations \& Broader Impact}
\label{app:limitations}

\textbf{Limitations:} \textsc{scShapeBench} has limitations that frame the scope of our claims. First, our shape vocabulary is deliberately restricted to the structures most commonly encountered in scRNA-seq analysis (clusters, trajectories, and bifurcations), and does not yet cover higher-dimensional features, nested structures, or cycles. Second, the expert-annotated corpus, while curated with care, reflects the judgment of a finite set of annotators and may encode biases in how ambiguous topologies are resolved. Finally, whether improvements on \textsc{scShapeBench} translate into better biological discoveries remains an open empirical question. Addressing these limitations, particularly expanding the shape vocabulary and connecting evaluation to downstream tasks, is a central direction for future work.

\textbf{Broader impact:} Beyond benchmarking, automating shape detection has the potential to make downstream analysis more effective. Practitioners default to clustering pipelines even when the data carries trajectory or archetypal structure, and biological signal is routinely lost as a result. A reliable, automated upstream step could surface this signal and, in agentic settings, allow AI systems to select analyses on the basis of data structure rather than convention. We believe \textsc{scShapeBench} provides a foundation for this work and a common ground on which new shape-detection methods can be developed and compared.

\section{Annotation Webpage}
\label{app:webpage}
\begin{figure}[H]
    \centering
    \includegraphics[width=.5\linewidth]{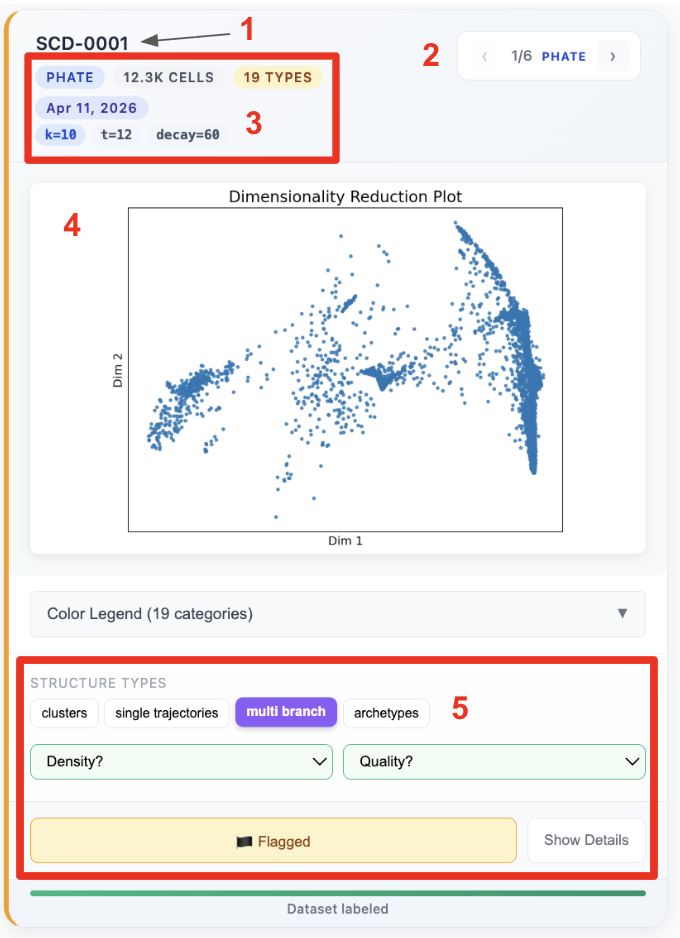}
    \caption{\textbf{Web interface for data annotation and visualization.} Key components include: \textbf{(1) Dataset Name} identifying the active dataset; \textbf{(2) Algorithm Selection} for switching between five PHATE plots (varying $k$) and a UMAP embedding; \textbf{(3) Visualization Parameters} displaying settings for the chosen algorithm; \textbf{(4) Visualization Plot} showing the point cloud embedding; and \textbf{(5) Labeling Tools} for non-exclusive labeling, density assessment, and data quality rating (poor, good, or excellent).}
    \label{fig:annotation_figure}
\end{figure}

\newpage
\section{Visualization of Example Graphs}
\label{app:visu_graphs}
This section provides representative examples of recovered graph structures on both the synthetic benchmark and real scRNA-seq datasets.

\begin{figure}[H]
    \centering
    \includegraphics[width=\linewidth]{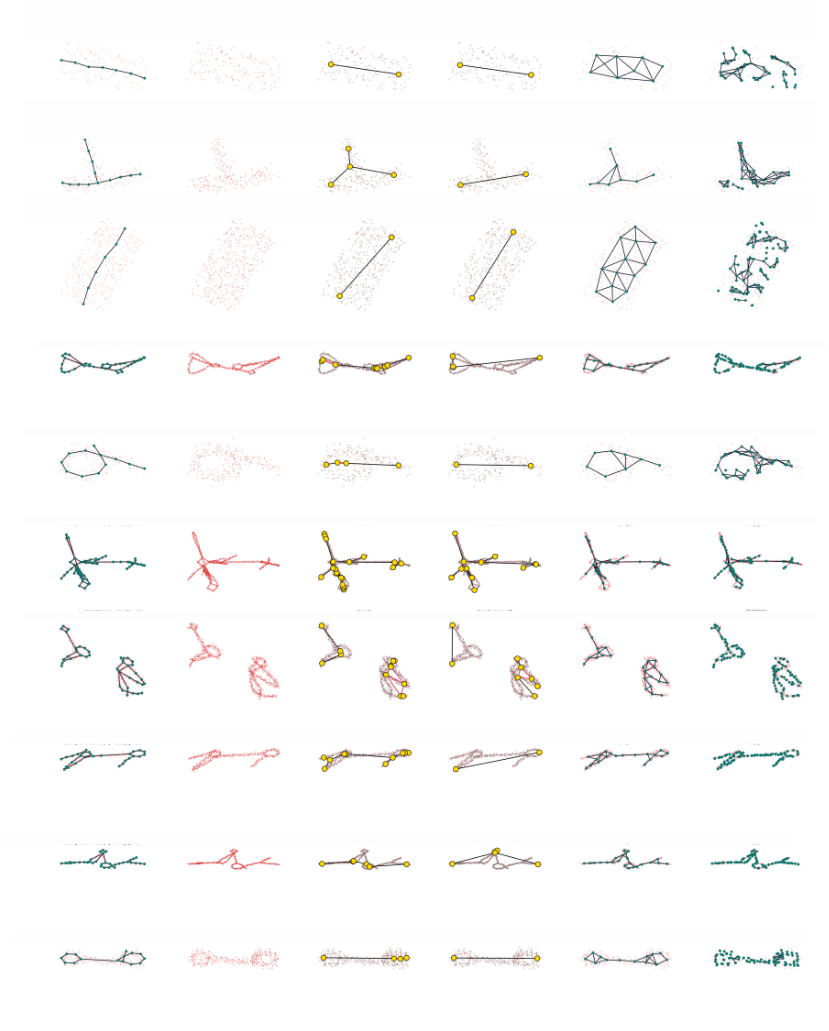}
    \caption{
    Representative synthetic examples comparing recovered graph structures across methods. From left to right: the latent synthetic graph overlaid on the sampled point cloud, the sampled point cloud alone, the first scale of \textsc{scReebTower} (\(\ell=0\)), a higher diffusion scale of \textsc{scReebTower}, PAGA, and Mapper. The examples illustrate recovery behavior across multiple topology classes and noise regimes.
    }
    \label{fig:synthetic_gallery}
\end{figure}

\begin{figure}[H]
    \centering
    \includegraphics[width=0.6\linewidth]{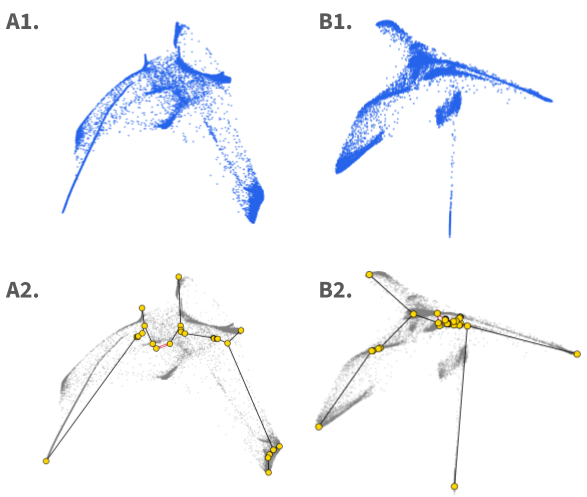}
    \caption{
    Representative real-world scRNA-seq datasets visualized using PHATE embeddings together with recovered \textsc{scReebTower} graphs. Panels A1 and B1 show the PHATE embeddings, while panels A2 and B2 show the corresponding recovered graph structures overlaid on the embeddings.
    }
    \label{fig:realworld_gallery}
\end{figure}

\end{document}